\documentclass[pdflatex,sn-mathphys-num]{sn-jnl}

\usepackage{graphicx}%
\usepackage{multirow}%
\usepackage{amsmath,amssymb,amsfonts}%
\usepackage{amsthm}%
\usepackage{mathrsfs}%
\usepackage[title]{appendix}%
\usepackage{xcolor}%
\usepackage{textcomp}%
\usepackage{manyfoot}%
\usepackage{booktabs}%
\usepackage{algorithm}%
\usepackage{algorithmicx}%
\usepackage{algpseudocode}%
\usepackage{listings}%
\usepackage{makecell}
\usepackage{float}

\theoremstyle{thmstyleone}%
\theoremstyle{thmstyletwo}%

\theoremstyle{thmstylethree}%

\begin{document}
\title[SurGrID]{SurGrID: Controllable \textit{Sur}gical Simulation via Scene \textit{Gr}aph to \textit{I}mage \textit{D}iffusion}


\author*[1]{\fnm{Yannik} \sur{Frisch}}\email{yannik.frisch@gris.tu-darmstadt.de}
\equalcont{These authors contributed equally to this work.}
\author*[1]{\fnm{Ssharvien} \sur{Kumar Sivakumar}}\email{ssharvien.kumar.sivakumar@gris.tu-darmstadt.de}
\equalcont{These authors contributed equally to this work.}
\author[2]{\fnm{\c{C}a\u{g}han} \sur{Köksal}}
\author[4]{\fnm{Elsa} \sur{Böhm}}
\author[4]{\fnm{Felix} \sur{Wagner}}
\author[4]{\fnm{Adrian} \sur{Gericke}}
\author[3]{\fnm{Ghazal} \sur{Ghazaei}}
\author[1]{\fnm{Anirban} \sur{Mukhopadhyay}}

\affil*[1]{\orgname{TU Darmstadt}, \orgaddress{\street{Fraunhoferstr. 5}, \city{Darmstadt}, \postcode{64297}, \country{Germany}}}
\affil[2]{\orgname{TU Munich}, \orgaddress{\street{Arcisstr. 21}, \city{Munich}, \postcode{80333}, \country{Germany}}}
\affil[3]{\orgname{Carl Zeiss AG}, \orgaddress{\street{Kistlerhofstr. 75}, \city{Munich}, \postcode{81379}, \country{Germany}}}
\affil[4]{\orgname{Universitätsmedizin Mainz}, \orgaddress{\street{Langenbeckstr. 1}, \city{Mainz}, \postcode{55131}, \country{Germany}}}


\abstract{
\textbf{Purpose:} Surgical simulation offers a promising addition to conventional surgical training. However, available simulation tools lack photorealism and rely on hard-coded behaviour. Denoising Diffusion Models are a promising alternative for high-fidelity image synthesis, but existing state-of-the-art conditioning methods fall short in providing precise control or interactivity over the generated scenes.

\textbf{Methods:} We introduce SurGrID, a Scene Graph to Image Diffusion Model, allowing for controllable surgical scene synthesis by leveraging Scene Graphs. These graphs encode a surgical scene's components' spatial and semantic information, which are then translated into an intermediate representation using our novel pre-training step that explicitly captures local and global information.

\textbf{Results:} Our proposed method improves the fidelity of generated images and their coherence with the graph input over the state-of-the-art. Further, we demonstrate the simulation's realism and controllability in a user assessment study involving clinical experts.

\textbf{Conclusion:} Scene Graphs can be effectively used for precise and interactive conditioning of Denoising Diffusion Models for simulating surgical scenes, enabling high fidelity and interactive control over the generated content. 
}




\maketitle

\section{Introduction}
\label{sec:intro}


The conventional methodology for training novice surgeons follows the Halstedian apprenticeship model, in which trainees are deemed competent after completing a minimum number of surgical procedures under supervision \cite{lee2020systematic}. Nevertheless, the Halstedian approach raises ethical concerns over using patients for training and has been associated with increased complication rates and poor outcomes \cite{kwong2014long}. However, when paired with simulation-based training, it offers a promising way to overcome regulatory and ethical complexities while establishing a safe and controlled environment where novice surgeons can refine their skills without the potentially severe consequences of real-world failures. It has also been shown that surgeons who have received simulation-based training demonstrate higher overall performance and make fewer errors during their initial surgeries than those who only received conventional training \cite{thomsen2017operating}.

In addition to using phantom or cadaver specimens for training simulation, numerous studies have shown the effectiveness of virtual reality \textbf{simulators for training} \cite{staropoli2018surgical, thomsen2017operating}. However, these tools are computer graphics-based, with manually defined rules for rendering logic. Therefore, new surgery techniques and edge cases must be manually programmed, but still fall short in replicating the complexity and variability of human anatomy and the subtle nuances of real-life surgical procedures \cite{iliash2024interactive}. Recent works propose using Denoising Diffusion Models (DDMs) for photorealistic interactive simulation to address it. However, these methods either rely on text prompts to guide generation \cite{cho2024surgen}, which offers limited spatial control, or use mask prompts \cite{iliash2024interactive}, which are not easily interactable and require continuous mask adjustments.

\begin{figure}[H]
    \centering
    \includegraphics[width=12.0cm]{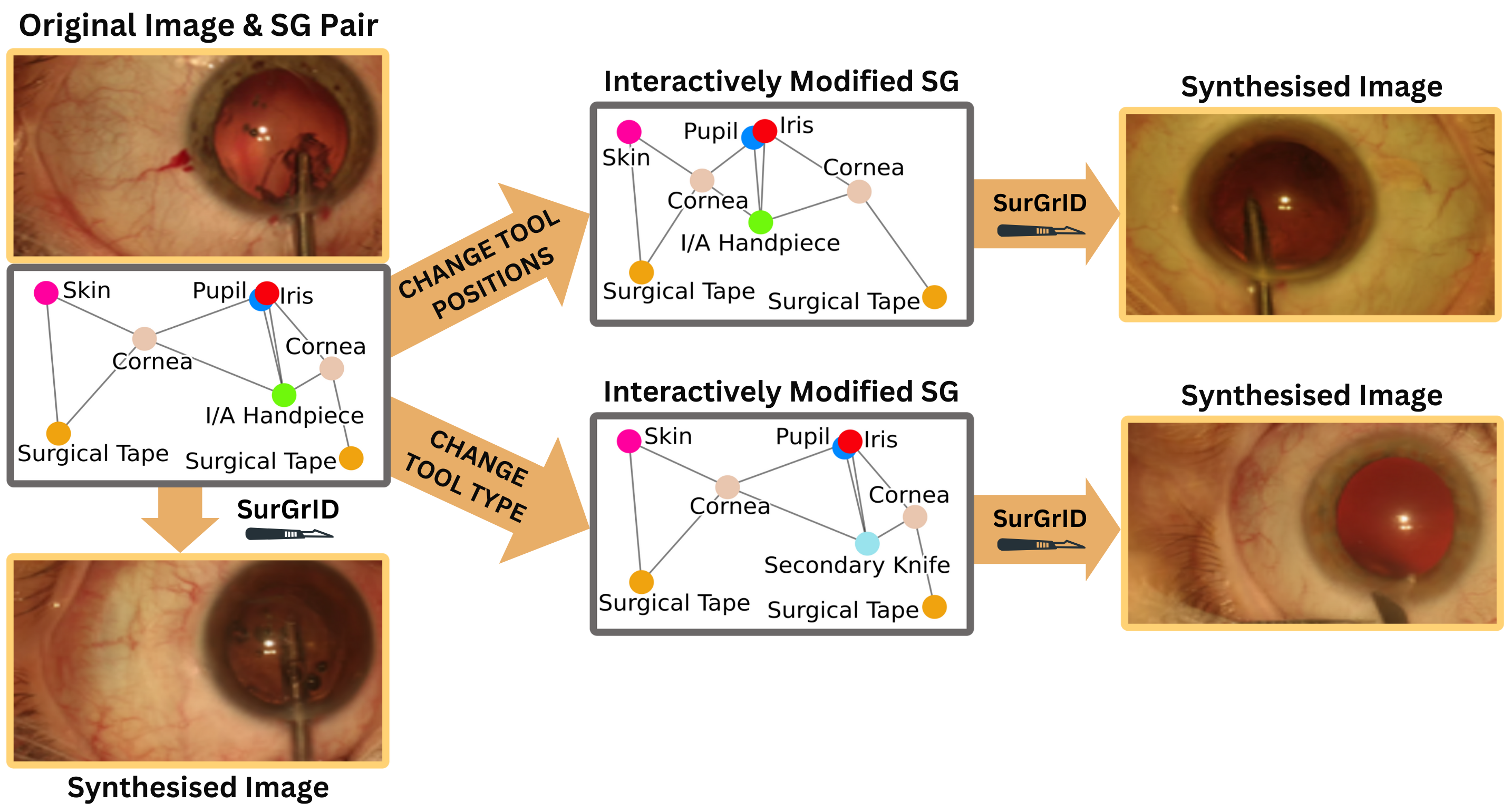}
    \caption{\textbf{Concept of SurGrID.} SurGrID conditions image synthesis on Scene Graphs (SGs) for precise control over the anatomy/tool type, size, and position. In the bottom left, SurGrID generates a new image spatially identical to the original image using only the ground-truth SG. This demonstrates that SGs can effectively encode a surgical scene's spatial and semantic information. On the right, we interactively modify the SG to control anatomy/tool position (top) and type (bottom).
    }
    \label{fig:intro}
\end{figure}

Recognizing this gap, we propose \textbf{SurGrID, a compact yet interpretable and precise conditioning of DDMs with Scene Graphs (SGs).} SurGrID is trained on surgical videos of actual patients and synthesises new surgical scenes with high controllability and excellent fidelity. At the core of our method lies the adoption of \textbf{SG representations as compact and human-readable encodings of surgical scenes} \cite{holm2023dynamic, murali2023latent, koksal2024sangria}. The hierarchical and relational nature of SGs makes them an ideal candidate for simulating dynamic and interactive surgical environments. We encode precise spatial information, such as the size and position of the anatomies and tools, into the node features of SGs obtained from a surgical scene, allowing fine-tuned control over the synthesis, as shown in Figure \ref{fig:intro}.

To synthesise images that accurately reflect the SG, we must acquire an intermediate representation that can align between the SG and images \cite{yang2022diffusion, mishra2024scene}. To this end, we propose \textbf{pre-training a graph encoder} on the \textbf{(image, segmentation, graph) triplet} that captures \textbf{local and global information} from the surgical scene. We focus on predicting fine-grained details of the anatomy and tools for local information by masking them randomly and reconstructing masked regions based on the SG input. We employ a discriminative approach to learn global information, which captures the layout and interaction between anatomy and tools. This approach enforces the encoder to cluster representations of compliant SGs and masks in the latent space.

We then leverage the obtained graph embeddings for \textbf{conditionally training a DDM} \cite{ho2020denoising} to synthesise realistic surgical images while allowing precise control over the generated content through graph conditioning. The ability to generate diverse and realistic images from such structured representations holds immense potential for surgical simulation, offering a novel avenue for creating varied and high-fidelity surgical scenes. To the best of our knowledge, this manuscript is the first to propose SG to image diffusion for precisely controllable surgical simulation.

\paragraph{Contributions}
\begin{itemize}
    \item We show that SGs can encode surgical scenes in a human-readable format. We propose a novel pre-training step that encodes global and local information from (image, mask, SG) triplets. The learned embeddings are employed to condition graph to image diffusion for high-quality and precisely controllable surgical simulation.
    
    \item We evaluate our generative approach on scenes from cataract surgeries using quantitative fidelity and diversity measurements, followed by an extensive user study involving clinical experts.
\end{itemize}

\section{Related Work}
\label{sec:rw}

Simulation by generative models typically conditions the synthesis on simple inputs such as class labels \cite{muller2023multimodal, frisch2023synthesising}, reference images \cite{kim2022diffusion, fuchsharp} or text queries \cite{allmendinger2024navigating}. However, none of these techniques provides a highly controllable and precise conditioning mechanism suitable for complex, high-risk scenarios such as surgical simulation. We postulate that Scene Graph conditioning can serve as an accurate conditioning mechanism.

Translating SG representations into realistic images has gained interest within the computer vision community. To achieve this task, Johnson et al. \cite{johnson2018image} use the intermediate representation of a \emph{scene layout}, which they predict from input SGs and translate into an image using a Cascaded Refinement Network. With the rise of Denoising Diffusion Models \cite{ho2020denoising,dhariwal2021diffusion,rombach2022high} as a competing generative method, Yang et al. \cite{yang2022diffusion} have laid foundational principles for SG to image synthesis with diffusion models, utilizing Masked Contrastive Pre-Training to obtain graph embeddings for conditioning. SceneGenie \cite{farshad2023scenegenie} guides the denoising process through information from CLIP embeddings \cite{radford2021learning} for input text queries together with scene layout and segmentation mask predictions from the text query. R3CD \cite{liu2024r3cd} introduces SG-Transformers to improve the global and local information in graph embeddings and uses a contrastive loss to improve relation-aware synthesis. Mishra et al. \cite{mishra2024scene} introduce adversarial pre-training of a graph encoder, aligning graph embeddings with CLIP image embeddings \cite{radford2021learning} and removing the need for an intermediate scene layout representation. 

These methods operate under the \textit{assumption that the image embeddings corresponding to a particular SG will be distant from those associated with entirely different SGs}. However, this assumption falls short in surgical simulation, rendering these approaches ineffective when applied directly. For instance, neighbouring frames in a surgical video may depict different tools and have distinct SGs. Yet, they occupy a similar position in the image embedding space due to similarly looking anatomy. As a result, frames from the same video tend to cluster together in the image embedding space. However, \textbf{we would like frames with similar surgical tools and positions to be close in the embedding space, regardless of the video they come from.} The existing methods further don’t address the need for precise control over positioning and sizing, which is essential for surgical simulation.

\section{Method}
\label{sec:meth}

This section presents our approach for translating Scene Graphs into intermediate representations that capture both local and global information. Subsequently, we outline the conditional diffusion model used for SG to image generation.

\subsection{Pre-Training Scene Graph Encoder}
We leverage a surgical segmentation dataset $\mathcal{D}$ consisting of (image, mask) pairs $(x, m) \in \mathcal{D}$ to \textbf{extract SGs from ground-truth segmentation masks}. Details on this conversion can be found in Supplementary Section \ref{sec:app_sg}. The approaches in subsequent sections can seamlessly be included in datasets with already available SGs.

For the graph encoder $E_G$, we employ a series of stacked Graph Neural Network (GNN) layers \cite{wu2020comprehensive} to process the input graph $\mathcal{G} = (\mathcal{V},\mathcal{E})$. Each GNN layer updates node representations by aggregating information from their neighbours, effectively capturing local graph topologies:

\begin{equation}
    h_v^{(l+1)} = \text{GNN}^{(l)}(h_v^{(l)}, \bigoplus_{u \in \mathcal{N}(v)} h_u^{(l)})
\end{equation}

where $h_v^{(l)}$ is the feature vector of node $v$ at layer $l, N(v)$ represents the neighbours of $v$, and $\bigoplus$ denotes a differentiable, permutation-invariant aggregation function, such as sum, mean, or max. The final node representations are aggregated through mean pooling to form a graph-level latent representation $z_G$. 

\begin{figure}[htbp]
    \centering
    \includegraphics[width=\linewidth]{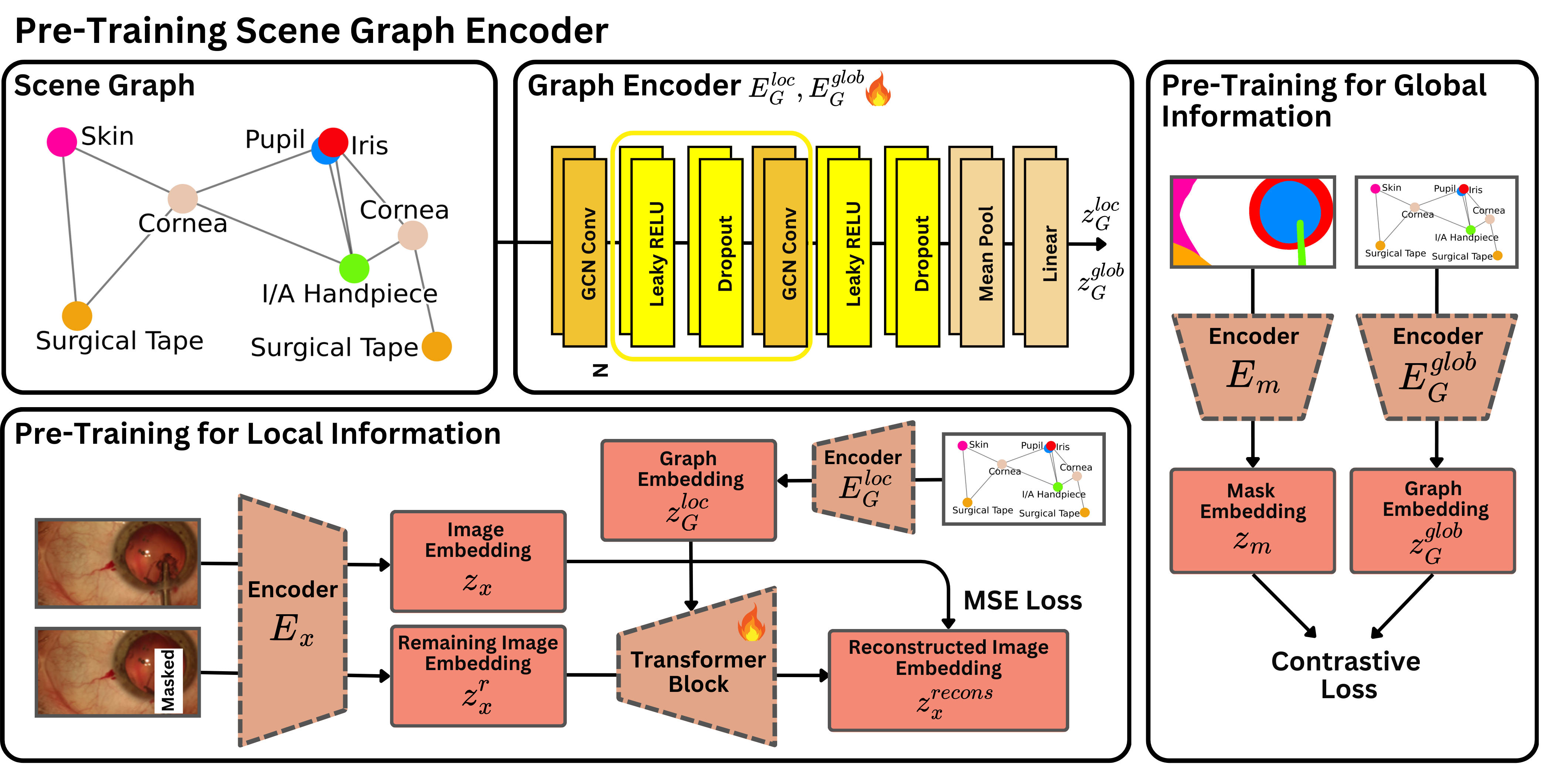}
    \caption{\textbf{SurGrID Workflow for Pre-training Graph Encoders} To capture local and global information, we pre-train two separate graph encoders composed of stacked GNN layers (top centre). $E_G^\text{loc}$ is trained to encode local information by randomly masking objects in the image and predicting the masked area using the SG (bottom). $E_G^\text{glob}$ is trained to encode global information by aligning the graph embedding space with the segmentation mask embedding space (right).}
    \label{fig:methodology}
\end{figure}

We propose pre-training two distinct graph encoders, as visualised in Figure \ref{fig:methodology}: $E_G^{loc}$, which focuses on capturing local information, hence embedding fine-grained details of the anatomy and surgical tools. $E_G^{glob}$ concentrates on retaining global information, focusing on the overall structural alignment based on compositions and interactions within the surgical scene. The overall pre-training occurs in the latent space, where embeddings for the image, $z_x$, and segmentation masks, $z_m$, are obtained using a VQ-GAN model \cite{esser2020taming}. These models are trained separately and referred to as $E_m$ and $E_x$ in Figure \ref{fig:methodology}. Supplementary Section \ref{sec:app_tsne} analyses and motivates the pre-training mechanisms using t-SNE visualisations of the embedding spaces.

For the \textbf{local information}, we randomly select a class from an image $x$'s paired mask and apply rectangular masking of that class's bounding box in the image. Both the full image $x$ and the masked image $x^r$ are encoded as $z_x = E_x(x)$ and $z_x^r = E_x(x^r)$. To train $E_G^{loc}$, the task is framed as predicting $z_x$ based on the available $z_x^r$ and it's corresponding SG representation $z_G^{loc} = E_G^{loc}(\mathcal{G})$. The graph encoder is trained in conjunction with a transformer-based decoder, $d_\theta$, to minimize the reconstruction loss:

\begin{equation}
    \mathcal{L}_{\text{local}} = \mathbb{E}_{(x, \mathcal{G}) \sim D} \left\| z_x - d_\theta \left( z_x^r, z_G^{loc} \right) \right\|_2^2
\end{equation}

\indent To capture the \textbf{global information} of the surgical scene, we train $E_G^{glob}$ by aligning its graph embeddings, $z_G^{glob} = E_G^{glob}(\mathcal{G})$, with the segmentation mask embeddings, $z_m$, inspired by CLIP's training approach \cite{radford2021learning}. This alignment must not be performed in the image embedding space, as neighbouring frames with different tools in a surgical video often share a large portion of the anatomical features, causing them to cluster in the embedding space. Our goal, however, is to ensure that frames featuring the same tool, even across different videos and possibly different textural features, are grouped in the embedding space. To achieve this, we include segmentation mask embeddings, as they capture high-level information about the overall scene layout and are a more detailed representation of the SG. This enables them to accurately represent the SG in the embedding space, bringing similar SGs closer together. For training, we employ a contrastive loss:

\begin{equation}
    \mathcal{L}_{\text{global}} = \mathbb{E}_{(\mathcal{G}, m^+, \{{m_i}^-\}^k_{i=1}) \sim D} \left[ -\log \frac{\exp \left( z_G \cdot z_{m^+} \right)}{\exp \left( z_G \cdot z_{m^+} \right) + \sum_i \exp \left( z_G \cdot z_{{m_i}^-} \right)} \right]
\end{equation}

where $m^+$ represents the compliant segmentation mask to $\mathcal{G}$, while ${m_i}^-$ are the negative targets, which do not comply with $\mathcal{G}$ and are randomly sampled from $\mathcal{D}$.

\subsection{SG-Conditioned Denoising Diffusion Model}
For image synthesis, we combine the standard diffusion model setup outlined in DDM \cite{ho2020denoising} with Classifier-Free Guidance (CFG) \cite{ho2022classifier}. The conditioning $c = concat(z_G^{loc}, z_G^{glob})$ is derived by fusing the embeddings from the two graph encoders $E_G^{loc}, E_G^{glob}$. We then train a parameterised denoising model $\epsilon_\theta$ to approximate denoising steps $p(x_{t-1} | x_t, c)$ by minimising:

\begin{equation}
    \mathcal{L}_{\text{DDM}} = \mathbb{E}_{t, x_0, \epsilon} \left[ \left\| \epsilon - \epsilon_\theta (x_t, t, c) \right\|^2 \right]
\end{equation}

where $x_t = \sqrt{\bar{\alpha}_t} x_0 + \sqrt{1 - \bar{\alpha}_t} \epsilon, \ \epsilon \sim \mathcal{N}(0, I)$, and $\bar{\alpha}_t$ is the variance schedule of the diffusion process. By then sampling $z_T$ from a Gaussian distribution $p(z_T) \sim \mathcal{N}(0, I)$, we can synthesise new samples by iteratively applying the denoising network. 

CFG substitutes the predicted noise $\epsilon_\theta(z_t, t, c)$ at each diffusion time-step $t$ as:

\begin{equation}
    \epsilon'_\theta(x_t, t, c) := (1 + \omega) \epsilon_\theta(x_t, t, c) - \omega \epsilon_\theta(x_t, t)
\end{equation}

where $\omega$ is a scaling parameter, which we empirically set to 2.0, and $\epsilon_\theta(x_t, t)$ is trained by randomly dropping the conditioning with probability 0.2.
\section{Experiments and Results}
\label{sec:exp}
This section outlines our experimental setup and provides quantitative and qualitative evaluations of the high-quality surgical samples generated from Scene Graphs. We conclude with a visual assessment user study involving clinical experts.

\subsection{Setup and Datasets}
We evaluate our approach on the \emph{CaDIS} dataset \cite{grammatikopoulou2021cadis}, a semantic segmentation dataset for cataract surgery videos. Cataract surgery is one of the most frequently performed procedures worldwide \cite{allen2006cataract}, where clouded lenses are removed and replaced with artificial ones. The dataset captures 4670 frames from 25 videos which are at least 0.3s apart. They have a resolution of $960\times540$ pixels. In setting II, as defined in \emph{CaDIS} \cite{grammatikopoulou2021cadis}, the dataset includes segmentation masks for 17 classes divided into four anatomical labels, ten tool labels, and three miscellaneous labels. 

We split the examples based on the 25 available videos, allocating 19 for training, 3 for validation, and 3 for testing, resulting in 3550, 534, and 586 (image, mask, graph) triplets, respectively. Our model was trained on an NVIDIA RTX 4090 GPU with an image size of $128\times128$. The SG representations, model weights and code to reproduce our results will be provided at \href{https://github.com/MECLabTUDA/SurGrID}{https://github.com/MECLabTUDA/SurGrID} upon acceptance.

\subsection{Quantitative Image Analysis}
To evaluate the objective image quality of generated samples, we synthesise examples from ground-truth SGs and assess the FID and KID scores \cite{binkowski2018demystifying} against the same amount of real samples. Further, we evaluate the diversity of generated samples using the LPIPS diversity metric \cite{zhang2018unreasonable}.

To also assess how effectively the synthesised images adhere to the conditioning provided by the SG, we employ the Mask R-CNN model \cite{he2017mask} pre-trained on \emph{CaDIS} \cite{grammatikopoulou2021cadis} for object detection, which predicts bounding boxes (BBs) along with class labels. This process essentially reverses the workflow by converting images back into SGs. By assessing the BBs with the IoU metric (at a 50\% threshold) and evaluating the predicted object classes using the F1 score (at 50\% IoU threshold), we can assess the accuracy of the synthesised anatomies and tools, along with their corresponding positions and sizes. 

As an additional baseline, we deploy a Latent Diffusion Model (LDM) \cite{rombach2022high} conditioned on CLIP embeddings \cite{radford2021learning} of NLP strings. These strings resemble a combination of triplets, which we construct from spatial relations such as "left of," "right of," "above," "below," "inside," and "surrounding". The strings are then embedded using the CLIP text encoder, and the LDM is conditionally trained on the resulting embeddings and otherwise analogously to the other methods. 

\begin{table}[htbp]
    \centering
    \caption{\textbf{Quantiative Fidelity and Diversity Results.}}
    \begin{tabular}{l|ccc|cc}
         \textbf{Method} & \textbf{FID} ($\downarrow$) & \textbf{KID} ($\downarrow$) & \textbf{LPIPS} ($\uparrow$) & \textbf{BB IoU$@0.5$} ($\uparrow$) & \textbf{F1$@0.5$} ($\uparrow$)\\
         \hline
         Real Image & - & - & 0.599 & 0.636 & 0.585\\
         \hline
         Sg2Im \cite{johnson2018image} & 94.9 & 0.096 & 0.416 & 0.305 & 0.197\\
         LDM \cite{rombach2022high} (CLIP-cond.) & 38.6 & 0.033 & 0.455 & 0.310 & 0.169\\ 
         SGDiff \cite{yang2022diffusion} ($\omega=2.0$)  & 42.0 &  0.033 &  0.449 & 0.372 & 0.224\\     
         SurGrID ($\omega=2.0$) & \textbf{26.6} & \textbf{0.019} & \textbf{0.456} & \textbf{0.549} & \textbf{0.424}\\
    \end{tabular}
    \label{tab:quan}
\end{table}

As presented in Table \ref{tab:quan}, our method not only surpasses the baselines regarding FID and KID scores, reflecting improved image quality but also adheres more closely to the conditioning criteria, as shown by the higher BB IoU and F1 scores.

\subsection{Ablation Studies}

We present ablation scores for our method trained exclusively on embeddings containing either only global or local information in the top part of Table \ref{tab:ablation}. Combining both embeddings significantly improves the method's performance, highlighting the necessity to include both types of information in the generative process. The bottom part of Table \ref{tab:ablation} additionally displays ablation results on the guidance scale $\omega$.

\begin{table}[htbp]
    \centering
    \caption{\textbf{Guidance Scale and Embedding Pre-Training Ablation.}}
    \begin{tabular}{l|ccc|cc}
         \textbf{Method} & \textbf{FID} ($\downarrow$) & \textbf{KID} ($\downarrow$) & \textbf{LPIPS} ($\uparrow$) & \textbf{BB IoU$@0.5$} ($\uparrow$) & \textbf{F1$@0.5$} ($\uparrow$)\\
         \hline
         Real Image & - & - & 0.599 & 0.636 & 0.585\\
         \hline
         SurGrID (Local only) & 93.2 & 0.083 & 0.313 & 0.247 & 0.169\\
         SurGrID (Global only) & 120.2 & 0.111 & 0.536 & 0.199 & 0.128\\
         SurGrID ($\omega=2.0$) & \textbf{26.6} & \textbf{0.019} & \textbf{0.456} & \textbf{0.549} & \textbf{0.424}\\
         \hline
         SGDiff \cite{yang2022diffusion} ($\omega=1.0$) & 41.7 & 0.033 & 0.449 & 0.368 & 0.225\\
         SGDiff \cite{yang2022diffusion} ($\omega=2.0$) & 42.0 & 0.033 & 0.449 & 0.372 & 0.224\\
         SGDiff \cite{yang2022diffusion} ($\omega=3.0$) & 42.1 & 0.034 & 0.449 & 0.368 & 0.223\\
         SGDiff \cite{yang2022diffusion} ($\omega=4.0$) & 43.0 & 0.036 & 0.448 & 0.366 & 0.222\\
         SGDiff \cite{yang2022diffusion} ($\omega=5.0$) & 42.2 & 0.033 & 0.449 & 0.367 & 0.222\\
         \hline
         SurGrID ($\omega=1.0$) & \textbf{26.1} & \textbf{0.019} & 0.455 & \textbf{0.551} & 0.423\\
         SurGrID ($\omega=2.0$) & 26.6 & \textbf{0.019} & 0.456 & 0.549 & \textbf{0.424}\\
         SurGrID ($\omega=3.0$) & 47.3 & 0.041 & 0.436 & 0.465 & 0.314\\
         SurGrID ($\omega=4.0$) & 81.7 & 0.074 & 0.452 & 0.348 & 0.203\\
         SurGrID ($\omega=5.0$) & 114.0 & 0.114 & \textbf{0.464} & 0.300 & 0.162\\
    \end{tabular}
    \label{tab:ablation}
\end{table}

\begin{figure}
    \centering
    \includegraphics[width=\linewidth]{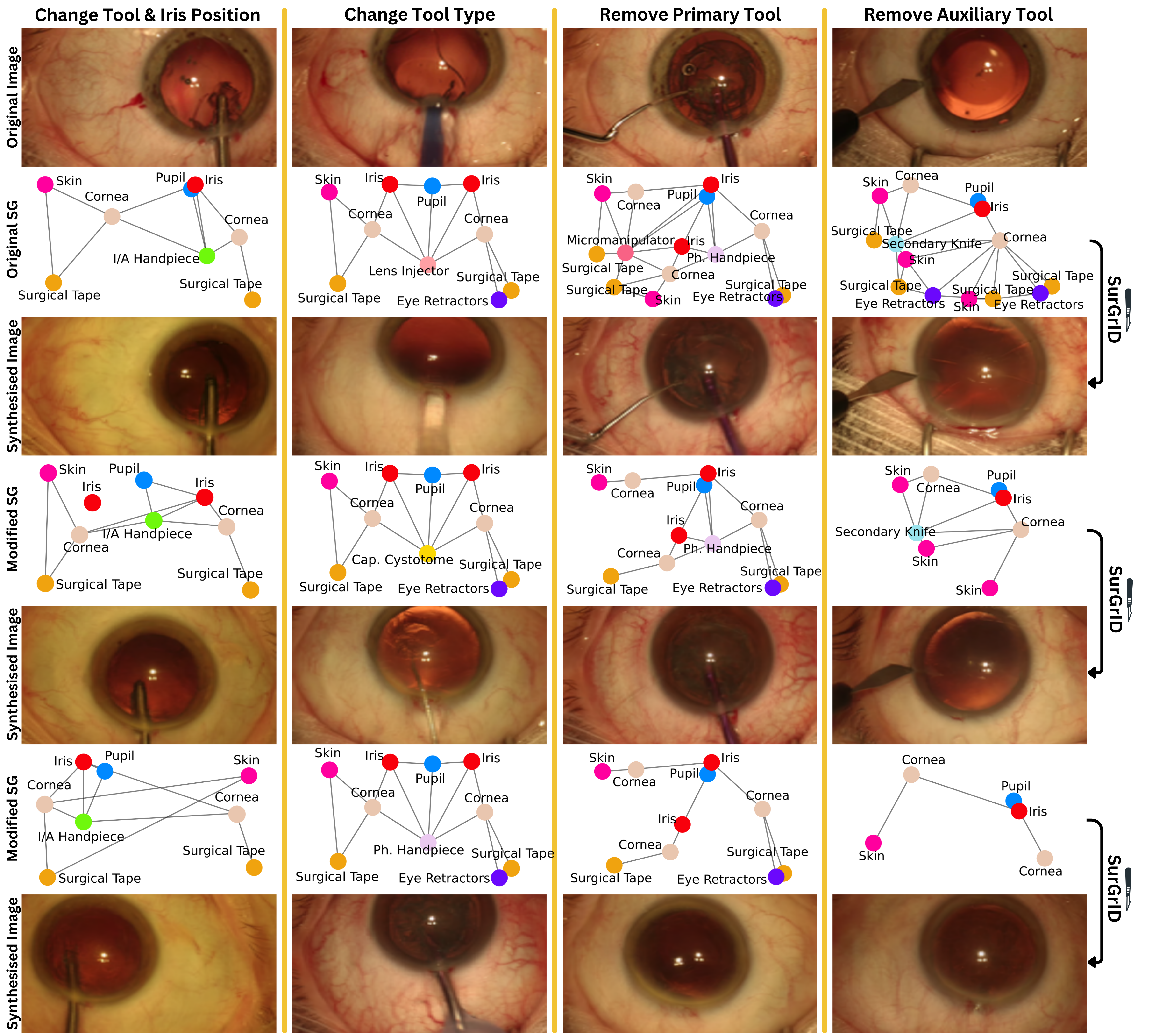}
    \caption{\textbf{Scene Graph To Image Generation.} The top row displays the original image, followed by the corresponding SG in the second row and synthesized images from the original SGs in the third row. Readers can judge the spatial coherence of the synthesised images in the third row by comparing them to the original image. Rows four and six present modified SGs; underneath them are synthesized images conditioned on these SGs.}
    \label{fig:qual}
\end{figure}

\subsection{Qualitative Results}
Besides quantitative evaluations, we visually assess the quality of our generated images and how well changes to the input graph translate to the synthesised images. As shown in Figure \ref{fig:qual} our generative model can synthesise high-quality images with coherent visual semantics following the SG inputs. 

The first column presents qualitative results from altering the positions of tools and anatomy within the SG, with the synthesised images accurately reflecting these positional adjustments. In the second column, the SG modifications focus solely on changing tool types. The generated images maintain a consistent spatial position while only adapting the type of displayed tool. Even though the tools can be displayed with different angles in the image, the centroid coordinates used for conditioning remain accurate. The third column progressively removes primary surgical tools, while the fourth column removes auxiliary tools such as retractors and surgical tape. In Appendix Section \ref{sec:app_lim}, we have outlined limitations and future work. However, in most cases, the synthesised images reliably show high fidelity and adhere to the input SG.

\subsection{Clinical Expert Assessment}
\label{sec:study}
We also conduct a visual assessment user study involving three ophthalmologists with at least two years of experience to evaluate the quality of our generative model. For simple and consistent statistical evaluation, we provide them with a graphical user interface (GUI) as displayed in Appendix Section \ref{sec:app_gui}. The GUI allows for loading ground-truth graphs along with the ground-truth image. The graph’s nodes can be moved, deleted, or have their class changed. We instruct our participants to load four different ground-truth graphs and sequentially perform the following actions on each: First, participants are instructed to generate a batch of four samples from the ground-truth SG without modifications. We request them to score the samples’ realism and coherence with the graph input using a Likert scale of 1 to 7. Here, 1 represents \emph{not realistic/coherent at all}, and 7 indicates \emph{totally realistic/coherent}. Second, the participants are requested to spatially move nodes in the canvas and again judge the synthesised samples. Third, participants change the class of one of the instrument nodes and judge the generated images. Lastly, participants are instructed to remove one of the instruments or miscellaneous classes and judge the synthesised image a final time. The study’s average results are summarised in Table \ref{tab:user}. Additionally, Section \ref{sec:app_fb} in the Appendix lists their summarised answers for the reasoning behind their ratings.

\begin{table}[htbp]
    \centering
    \caption{\textbf{Clinical Expert Assessment Study.} The abbreviation "Real." refers to the realism of the synthesised image, while "Coh." refers to the coherence with the input condition.}
    \label{tab:user}
    \renewcommand{\arraystretch}{1.1}
    \begin{tabular}{l|cc|cc|cc|cc} 
        & \multicolumn{2}{c|}{\textbf{Synth. from GT}} & \multicolumn{2}{c|}{\textbf{Spatial Modif.}} & \multicolumn{2}{c|}{\textbf{Tool Modif.}} & \multicolumn{2}{c}{\textbf{Tool Remov.}}
        \\
        \thead{\textbf{User}} & \thead{\textbf{Real.}} & \thead{\textbf{Coh.}} & \thead{\textbf{Real.}} & \thead{\textbf{Coh.}} & \thead{\textbf{Real.}} & \thead{\textbf{Coh.}} & \thead{\textbf{Real.}} & \thead{\textbf{Coh.}} \\
        \hline
        \textbf{P1} & 6.5$\pm$0.5 & 6.5$\pm$1.0 & 6.3$\pm$0.9 & 6.3$\pm$0.9 & 5.3$\pm$1.2 & 4.5$\pm$1.9 & 6.3$\pm$0.9 & 5.5$\pm$2.3 \\
        \textbf{P2} & 5.3$\pm$0.9 & 5.3$\pm$0.5 & 4.5$\pm$0.5 & 4.3$\pm$2.0 & 5.3$\pm$0.9 & 5.8$\pm$0.9 & 5.5$\pm$1.2 & 5.5$\pm$1.9 \\
        \textbf{P3} & 6.3$\pm$0.9 & 6.3$\pm$0.9 & 6.5$\pm$1.0 & 5.5$\pm$0.5 & 6.0$\pm$0.8 & 6.8$\pm$0.5 & 6.3$\pm$0.5 & 6.5$\pm$0.5 \\

        \hline
        \textbf{Average} & 6.0$\pm$0.9 & 6.0$\pm$0.9 & 5.8$\pm$1.2 & 5.3$\pm$1.4 & 5.5$\pm$1.0 & 5.7$\pm$1.4 & 6.0$\pm$0.9 & 5.8$\pm$1.6 \\
    \end{tabular}
\end{table}

On average, participants took around 30 minutes to complete the user study. They were allowed unlimited time to assess the synthesised images, enabling them to thoroughly inspect for any subtle artefacts that might not be immediately noticeable. Despite the extended review time, participants consistently found the synthesised images realistic, with an average score of 5.82, and coherent, with an average score of 5.70, with the changes they had made to the SG.
\section{Conclusions}
\label{sec:concl}
We present \textbf{Controllable Surgical Simulation via Scene Graph to Image Diffusion (SurGrID)}, the first controllable surgical simulator based on Scene Graph to Image Diffusion. We demonstrate the potential of SGs to encode semantic and spatial information of surgical scenes and as informative conditioning for synthesising new, unseen images with Denoising Diffusion Models. We also demonstrate that by interactively modifying the SGs, changes are directly reflected in the generated image, enabling precise control over the generation process. This overcomes the limitations of text prompts, which lack precision, and segmentation masks, which are difficult to modify and interact with. In our user study, surgeons verified the generated images to be realistic and coherent with the changes made to the scene graphs, highlighting SurGrID's substantial potential for realistic surgical simulation controlled by scene graphs. Our method surpasses state-of-the-art techniques in image quality and coherence to the graph input, which we demonstrate quantitatively and qualitatively. This paves the way for photorealistic surgical simulations that are trainable from real surgical videos while retaining the high controllability needed for surgical simulation. 

\backmatter

\bmhead{Supplementary information}

The supplementary information comprises the Appendix of the main manuscript.

\section*{Declarations}

\textbf{Funding.} This work has been partially funded by the German Federal Ministry of Education and Research as part of the Software Campus programme (project 500 01 528). 
\\
\textbf{Data Availability.} All experiments were conducted on publicly available datasets.
\\
\textbf{Code Availability.} Code will be published upon acceptance.
\\
Other declarations are not applicable.

\noindent

\bibliography{sn-bibliography}


\begin{thebibliography}{31}
\ifx \bisbn   \undefined \def \bisbn  #1{ISBN #1}\fi
\ifx \binits  \undefined \def \binits#1{#1}\fi
\ifx \bauthor  \undefined \def \bauthor#1{#1}\fi
\ifx \batitle  \undefined \def \batitle#1{#1}\fi
\ifx \bjtitle  \undefined \def \bjtitle#1{#1}\fi
\ifx \bvolume  \undefined \def \bvolume#1{\textbf{#1}}\fi
\ifx \byear  \undefined \def \byear#1{#1}\fi
\ifx \bissue  \undefined \def \bissue#1{#1}\fi
\ifx \bfpage  \undefined \def \bfpage#1{#1}\fi
\ifx \blpage  \undefined \def \blpage #1{#1}\fi
\ifx \burl  \undefined \def \burl#1{\textsf{#1}}\fi
\ifx \doiurl  \undefined \def \doiurl#1{\url{https://doi.org/#1}}\fi
\ifx \betal  \undefined \def \betal{\textit{et al.}}\fi
\ifx \binstitute  \undefined \def \binstitute#1{#1}\fi
\ifx \binstitutionaled  \undefined \def \binstitutionaled#1{#1}\fi
\ifx \bctitle  \undefined \def \bctitle#1{#1}\fi
\ifx \beditor  \undefined \def \beditor#1{#1}\fi
\ifx \bpublisher  \undefined \def \bpublisher#1{#1}\fi
\ifx \bbtitle  \undefined \def \bbtitle#1{#1}\fi
\ifx \bedition  \undefined \def \bedition#1{#1}\fi
\ifx \bseriesno  \undefined \def \bseriesno#1{#1}\fi
\ifx \blocation  \undefined \def \blocation#1{#1}\fi
\ifx \bsertitle  \undefined \def \bsertitle#1{#1}\fi
\ifx \bsnm \undefined \def \bsnm#1{#1}\fi
\ifx \bsuffix \undefined \def \bsuffix#1{#1}\fi
\ifx \bparticle \undefined \def \bparticle#1{#1}\fi
\ifx \barticle \undefined \def \barticle#1{#1}\fi
\bibcommenthead
\ifx \bconfdate \undefined \def \bconfdate #1{#1}\fi
\ifx \botherref \undefined \def \botherref #1{#1}\fi
\ifx \url \undefined \def \url#1{\textsf{#1}}\fi
\ifx \bchapter \undefined \def \bchapter#1{#1}\fi
\ifx \bbook \undefined \def \bbook#1{#1}\fi
\ifx \bcomment \undefined \def \bcomment#1{#1}\fi
\ifx \oauthor \undefined \def \oauthor#1{#1}\fi
\ifx \citeauthoryear \undefined \def \citeauthoryear#1{#1}\fi
\ifx \endbibitem  \undefined \def \endbibitem {}\fi
\ifx \bconflocation  \undefined \def \bconflocation#1{#1}\fi
\ifx \arxivurl  \undefined \def \arxivurl#1{\textsf{#1}}\fi
\csname PreBibitemsHook\endcsname

\bibitem[\protect\citeauthoryear{Lee et~al.}{2020}]{lee2020systematic}
\begin{barticle}
\bauthor{\bsnm{Lee}, \binits{R.}},
\bauthor{\bsnm{Raison}, \binits{N.}},
\bauthor{\bsnm{Lau}, \binits{W.Y.}},
\bauthor{\bsnm{Aydin}, \binits{A.}},
\bauthor{\bsnm{Dasgupta}, \binits{P.}},
\bauthor{\bsnm{Ahmed}, \binits{K.}},
\bauthor{\bsnm{Haldar}, \binits{S.}}:
\batitle{A systematic review of simulation-based training tools for technical
  and non-technical skills in ophthalmology}.
\bjtitle{Eye}
\bvolume{34}(\bissue{10}),
\bfpage{1737}--\blpage{1759}
(\byear{2020})
\end{barticle}
\endbibitem

\bibitem[\protect\citeauthoryear{Kwong et~al.}{2014}]{kwong2014long}
\begin{barticle}
\bauthor{\bsnm{Kwong}, \binits{A.}},
\bauthor{\bsnm{Law}, \binits{S.K.}},
\bauthor{\bsnm{Kule}, \binits{R.R.}},
\bauthor{\bsnm{Nouri-Mahdavi}, \binits{K.}},
\bauthor{\bsnm{Coleman}, \binits{A.L.}},
\bauthor{\bsnm{Caprioli}, \binits{J.}},
\bauthor{\bsnm{Giaconi}, \binits{J.A.}}:
\batitle{Long-term outcomes of resident-versus attending-performed primary
  trabeculectomy with mitomycin c in a united states residency program}.
\bjtitle{American journal of ophthalmology}
\bvolume{157}(\bissue{6}),
\bfpage{1190}--\blpage{1201}
(\byear{2014})
\end{barticle}
\endbibitem

\bibitem[\protect\citeauthoryear{Thomsen et~al.}{2017}]{thomsen2017operating}
\begin{barticle}
\bauthor{\bsnm{Thomsen}, \binits{A.S.S.}},
\bauthor{\bsnm{Bach-Holm}, \binits{D.}},
\bauthor{\bsnm{Kj{\ae}rbo}, \binits{H.}},
\bauthor{\bsnm{H{\o}jgaard-Olsen}, \binits{K.}},
\bauthor{\bsnm{Subhi}, \binits{Y.}},
\bauthor{\bsnm{Saleh}, \binits{G.M.}},
\bauthor{\bsnm{Park}, \binits{Y.S.}},
\bauthor{\bsnm{La~Cour}, \binits{M.}},
\bauthor{\bsnm{Konge}, \binits{L.}}:
\batitle{Operating room performance improves after proficiency-based virtual
  reality cataract surgery training}.
\bjtitle{Ophthalmology}
\bvolume{124}(\bissue{4}),
\bfpage{524}--\blpage{531}
(\byear{2017})
\end{barticle}
\endbibitem

\bibitem[\protect\citeauthoryear{Staropoli
  et~al.}{2018}]{staropoli2018surgical}
\begin{barticle}
\bauthor{\bsnm{Staropoli}, \binits{P.C.}},
\bauthor{\bsnm{Gregori}, \binits{N.Z.}},
\bauthor{\bsnm{Junk}, \binits{A.K.}},
\bauthor{\bsnm{Galor}, \binits{A.}},
\bauthor{\bsnm{Goldhardt}, \binits{R.}},
\bauthor{\bsnm{Goldhagen}, \binits{B.E.}},
\bauthor{\bsnm{Shi}, \binits{W.}},
\bauthor{\bsnm{Feuer}, \binits{W.}}:
\batitle{Surgical simulation training reduces intraoperative cataract surgery
  complications among residents}.
\bjtitle{Simulation in Healthcare}
\bvolume{13}(\bissue{1}),
\bfpage{11}--\blpage{15}
(\byear{2018})
\end{barticle}
\endbibitem

\bibitem[\protect\citeauthoryear{Iliash et~al.}{2024}]{iliash2024interactive}
\begin{bchapter}
\bauthor{\bsnm{Iliash}, \binits{I.}},
\bauthor{\bsnm{Allmendinger}, \binits{S.}},
\bauthor{\bsnm{Meissen}, \binits{F.}},
\bauthor{\bsnm{K{\"u}hl}, \binits{N.}},
\bauthor{\bsnm{R{\"u}ckert}, \binits{D.}}:
\bctitle{Interactive generation of laparoscopic videos with diffusion models}.
In: \bbtitle{MICCAI Workshop on Deep Generative Models},
pp. \bfpage{109}--\blpage{118}
(\byear{2024}).
\bcomment{Springer}
\end{bchapter}
\endbibitem

\bibitem[\protect\citeauthoryear{Cho et~al.}{2024}]{cho2024surgen}
\begin{botherref}
\oauthor{\bsnm{Cho}, \binits{J.}},
\oauthor{\bsnm{Schmidgall}, \binits{S.}},
\oauthor{\bsnm{Zakka}, \binits{C.}},
\oauthor{\bsnm{Mathur}, \binits{M.}},
\oauthor{\bsnm{Shad}, \binits{R.}},
\oauthor{\bsnm{Hiesinger}, \binits{W.}}:
Surgen: Text-guided diffusion model for surgical video generation.
arXiv preprint arXiv:2408.14028
(2024)
\end{botherref}
\endbibitem

\bibitem[\protect\citeauthoryear{Holm et~al.}{2023}]{holm2023dynamic}
\begin{bchapter}
\bauthor{\bsnm{Holm}, \binits{F.}},
\bauthor{\bsnm{Ghazaei}, \binits{G.}},
\bauthor{\bsnm{Czempiel}, \binits{T.}},
\bauthor{\bsnm{{\"O}zsoy}, \binits{E.}},
\bauthor{\bsnm{Saur}, \binits{S.}},
\bauthor{\bsnm{Navab}, \binits{N.}}:
\bctitle{Dynamic scene graph representation for surgical video}.
In: \bbtitle{ICCV Workshop},
pp. \bfpage{81}--\blpage{87}
(\byear{2023})
\end{bchapter}
\endbibitem

\bibitem[\protect\citeauthoryear{Murali et~al.}{2023}]{murali2023latent}
\begin{botherref}
\oauthor{\bsnm{Murali}, \binits{A.}},
\oauthor{\bsnm{Alapatt}, \binits{D.}},
\oauthor{\bsnm{Mascagni}, \binits{P.}},
\oauthor{\bsnm{Vardazaryan}, \binits{A.}},
\oauthor{\bsnm{Garcia}, \binits{A.}},
\oauthor{\bsnm{Okamoto}, \binits{N.}},
\oauthor{\bsnm{Mutter}, \binits{D.}},
\oauthor{\bsnm{Padoy}, \binits{N.}}:
Latent graph representations for critical view of safety assessment.
IEEE Transactions on Medical Imaging
(2023)
\end{botherref}
\endbibitem

\bibitem[\protect\citeauthoryear{K{\"o}ksal et~al.}{2024}]{koksal2024sangria}
\begin{botherref}
\oauthor{\bsnm{K{\"o}ksal}, \binits{{\c{C}}.}},
\oauthor{\bsnm{Ghazaei}, \binits{G.}},
\oauthor{\bsnm{Holm}, \binits{F.}},
\oauthor{\bsnm{Farshad}, \binits{A.}},
\oauthor{\bsnm{Navab}, \binits{N.}}:
Sangria: Surgical video scene graph optimization for surgical workflow
  prediction.
arXiv preprint arXiv:2407.20214
(2024)
\end{botherref}
\endbibitem

\bibitem[\protect\citeauthoryear{Yang et~al.}{2022}]{yang2022diffusion}
\begin{botherref}
\oauthor{\bsnm{Yang}, \binits{L.}},
\oauthor{\bsnm{Huang}, \binits{Z.}},
\oauthor{\bsnm{Song}, \binits{Y.}},
\oauthor{\bsnm{Hong}, \binits{S.}},
\oauthor{\bsnm{Li}, \binits{G.}},
\oauthor{\bsnm{Zhang}, \binits{W.}},
\oauthor{\bsnm{Cui}, \binits{B.}},
\oauthor{\bsnm{Ghanem}, \binits{B.}},
\oauthor{\bsnm{Yang}, \binits{M.-H.}}:
Diffusion-based scene graph to image generation with masked contrastive
  pre-training.
arXiv preprint arXiv:2211.11138
(2022)
\end{botherref}
\endbibitem

\bibitem[\protect\citeauthoryear{Mishra and
  Subramanyam}{2024}]{mishra2024scene}
\begin{botherref}
\oauthor{\bsnm{Mishra}, \binits{R.}},
\oauthor{\bsnm{Subramanyam}, \binits{A.}}:
Scene graph to image synthesis: Integrating clip guidance with graph
  conditioning in diffusion models.
arXiv preprint arXiv:2401.14111
(2024)
\end{botherref}
\endbibitem

\bibitem[\protect\citeauthoryear{Ho et~al.}{2020}]{ho2020denoising}
\begin{barticle}
\bauthor{\bsnm{Ho}, \binits{J.}},
\bauthor{\bsnm{Jain}, \binits{A.}},
\bauthor{\bsnm{Abbeel}, \binits{P.}}:
\batitle{Denoising diffusion probabilistic models}.
\bjtitle{NeurIPS}
\bvolume{33},
\bfpage{6840}--\blpage{6851}
(\byear{2020})
\end{barticle}
\endbibitem

\bibitem[\protect\citeauthoryear{M{\"u}ller-Franzes
  et~al.}{2023}]{muller2023multimodal}
\begin{barticle}
\bauthor{\bsnm{M{\"u}ller-Franzes}, \binits{G.}},
\bauthor{\bsnm{Niehues}, \binits{J.M.}},
\bauthor{\bsnm{Khader}, \binits{F.}},
\bauthor{\bsnm{Arasteh}, \binits{S.T.}},
\bauthor{\bsnm{Haarburger}, \binits{C.}},
\bauthor{\bsnm{Kuhl}, \binits{C.}},
\bauthor{\bsnm{Wang}, \binits{T.}},
\bauthor{\bsnm{Han}, \binits{T.}},
\bauthor{\bsnm{Nolte}, \binits{T.}},
\bauthor{\bsnm{Nebelung}, \binits{S.}}, \betal:
\batitle{A multimodal comparison of latent denoising diffusion probabilistic
  models and generative adversarial networks for medical image synthesis}.
\bjtitle{Scientific Reports}
\bvolume{13}(\bissue{1}),
\bfpage{12098}
(\byear{2023})
\end{barticle}
\endbibitem

\bibitem[\protect\citeauthoryear{Frisch et~al.}{2023}]{frisch2023synthesising}
\begin{bchapter}
\bauthor{\bsnm{Frisch}, \binits{Y.}},
\bauthor{\bsnm{Fuchs}, \binits{M.}},
\bauthor{\bsnm{Sanner}, \binits{A.}},
\bauthor{\bsnm{Ucar}, \binits{F.A.}},
\bauthor{\bsnm{Frenzel}, \binits{M.}},
\bauthor{\bsnm{Wasielica-Poslednik}, \binits{J.}},
\bauthor{\bsnm{Gericke}, \binits{A.}},
\bauthor{\bsnm{Wagner}, \binits{F.M.}},
\bauthor{\bsnm{Dratsch}, \binits{T.}},
\bauthor{\bsnm{Mukhopadhyay}, \binits{A.}}:
\bctitle{Synthesising rare cataract surgery samples with guided diffusion
  models}.
In: \bbtitle{MICCAI},
pp. \bfpage{354}--\blpage{364}
(\byear{2023}).
\bcomment{Springer}
\end{bchapter}
\endbibitem

\bibitem[\protect\citeauthoryear{Kim and Ye}{2022}]{kim2022diffusion}
\begin{bchapter}
\bauthor{\bsnm{Kim}, \binits{B.}},
\bauthor{\bsnm{Ye}, \binits{J.C.}}:
\bctitle{Diffusion deformable model for 4d temporal medical image generation}.
In: \bbtitle{MICCAI},
pp. \bfpage{539}--\blpage{548}
(\byear{2022}).
\bcomment{Springer}
\end{bchapter}
\endbibitem

\bibitem[\protect\citeauthoryear{Fuchs et~al.}{}]{fuchsharp}
\begin{botherref}
\oauthor{\bsnm{Fuchs}, \binits{M.}},
\oauthor{\bsnm{Sivakumar}, \binits{S.K.R.}},
\oauthor{\bsnm{Sch{\"o}ber}, \binits{M.}},
\oauthor{\bsnm{Woltering}, \binits{N.}},
\oauthor{\bsnm{Eich}, \binits{M.-L.}},
\oauthor{\bsnm{Schweizer}, \binits{L.}},
\oauthor{\bsnm{Mukhopadhyay}, \binits{A.}}:
Harp: Unsupervised histopathology artifact restoration.
In: MIDL
\end{botherref}
\endbibitem

\bibitem[\protect\citeauthoryear{Allmendinger
  et~al.}{2024}]{allmendinger2024navigating}
\begin{botherref}
\oauthor{\bsnm{Allmendinger}, \binits{S.}},
\oauthor{\bsnm{Hemmer}, \binits{P.}},
\oauthor{\bsnm{Queisner}, \binits{M.}},
\oauthor{\bsnm{Sauer}, \binits{I.}},
\oauthor{\bsnm{M{\"u}ller}, \binits{L.}},
\oauthor{\bsnm{Jakubik}, \binits{J.}},
\oauthor{\bsnm{V{\"o}ssing}, \binits{M.}},
\oauthor{\bsnm{K{\"u}hl}, \binits{N.}}:
Navigating the synthetic realm: Harnessing diffusion-based models for
  laparoscopic text-to-image generation.
In: AI for Health Equity and Fairness: Leveraging AI to Address Social
  Determinants of Health,
Springer Nature Switzerland,
pp. 31--46
(2024)
\end{botherref}
\endbibitem

\bibitem[\protect\citeauthoryear{Johnson et~al.}{2018}]{johnson2018image}
\begin{bchapter}
\bauthor{\bsnm{Johnson}, \binits{J.}},
\bauthor{\bsnm{Gupta}, \binits{A.}},
\bauthor{\bsnm{Fei-Fei}, \binits{L.}}:
\bctitle{Image generation from scene graphs}.
In: \bbtitle{CVPR},
pp. \bfpage{1219}--\blpage{1228}
(\byear{2018})
\end{bchapter}
\endbibitem

\bibitem[\protect\citeauthoryear{Dhariwal and
  Nichol}{2021}]{dhariwal2021diffusion}
\begin{barticle}
\bauthor{\bsnm{Dhariwal}, \binits{P.}},
\bauthor{\bsnm{Nichol}, \binits{A.}}:
\batitle{Diffusion models beat gans on image synthesis}.
\bjtitle{Advances in Neural Information Processing Systems}
\bvolume{34},
\bfpage{8780}--\blpage{8794}
(\byear{2021})
\end{barticle}
\endbibitem

\bibitem[\protect\citeauthoryear{Rombach et~al.}{2022}]{rombach2022high}
\begin{bchapter}
\bauthor{\bsnm{Rombach}, \binits{R.}},
\bauthor{\bsnm{Blattmann}, \binits{A.}},
\bauthor{\bsnm{Lorenz}, \binits{D.}},
\bauthor{\bsnm{Esser}, \binits{P.}},
\bauthor{\bsnm{Ommer}, \binits{B.}}:
\bctitle{High-resolution image synthesis with latent diffusion models}.
In: \bbtitle{CVPR},
pp. \bfpage{10684}--\blpage{10695}
(\byear{2022})
\end{bchapter}
\endbibitem

\bibitem[\protect\citeauthoryear{Farshad et~al.}{2023}]{farshad2023scenegenie}
\begin{bchapter}
\bauthor{\bsnm{Farshad}, \binits{A.}},
\bauthor{\bsnm{Yeganeh}, \binits{Y.}},
\bauthor{\bsnm{Chi}, \binits{Y.}},
\bauthor{\bsnm{Shen}, \binits{C.}},
\bauthor{\bsnm{Ommer}, \binits{B.}},
\bauthor{\bsnm{Navab}, \binits{N.}}:
\bctitle{Scenegenie: Scene graph guided diffusion models for image synthesis}.
In: \bbtitle{ICCV Workshop},
pp. \bfpage{88}--\blpage{98}
(\byear{2023})
\end{bchapter}
\endbibitem

\bibitem[\protect\citeauthoryear{Radford et~al.}{2021}]{radford2021learning}
\begin{bchapter}
\bauthor{\bsnm{Radford}, \binits{A.}},
\bauthor{\bsnm{Kim}, \binits{J.W.}},
\bauthor{\bsnm{Hallacy}, \binits{C.}},
\bauthor{\bsnm{Ramesh}, \binits{A.}},
\bauthor{\bsnm{Goh}, \binits{G.}},
\bauthor{\bsnm{Agarwal}, \binits{S.}},
\bauthor{\bsnm{Sastry}, \binits{G.}},
\bauthor{\bsnm{Askell}, \binits{A.}},
\bauthor{\bsnm{Mishkin}, \binits{P.}},
\bauthor{\bsnm{Clark}, \binits{J.}}, \betal:
\bctitle{Learning transferable visual models from natural language
  supervision}.
In: \bbtitle{International Conference on Machine Learning},
pp. \bfpage{8748}--\blpage{8763}
(\byear{2021}).
\bcomment{PMLR}
\end{bchapter}
\endbibitem

\bibitem[\protect\citeauthoryear{Liu and Liu}{2024}]{liu2024r3cd}
\begin{bchapter}
\bauthor{\bsnm{Liu}, \binits{J.}},
\bauthor{\bsnm{Liu}, \binits{Q.}}:
\bctitle{R3cd: Scene graph to image generation with relation-aware
  compositional contrastive control diffusion}.
In: \bbtitle{AAAI},
vol. \bseriesno{38},
pp. \bfpage{3657}--\blpage{3665}
(\byear{2024})
\end{bchapter}
\endbibitem

\bibitem[\protect\citeauthoryear{Wu et~al.}{2020}]{wu2020comprehensive}
\begin{barticle}
\bauthor{\bsnm{Wu}, \binits{Z.}},
\bauthor{\bsnm{Pan}, \binits{S.}},
\bauthor{\bsnm{Chen}, \binits{F.}},
\bauthor{\bsnm{Long}, \binits{G.}},
\bauthor{\bsnm{Zhang}, \binits{C.}},
\bauthor{\bsnm{Philip}, \binits{S.Y.}}:
\batitle{A comprehensive survey on graph neural networks}.
\bjtitle{IEEE transactions on neural networks and learning systems}
\bvolume{32}(\bissue{1}),
\bfpage{4}--\blpage{24}
(\byear{2020})
\end{barticle}
\endbibitem

\bibitem[\protect\citeauthoryear{Esser et~al.}{2020}]{esser2020taming}
\begin{bchapter}
\bauthor{\bsnm{Esser}, \binits{P.}},
\bauthor{\bsnm{Rombach}, \binits{R.}},
\bauthor{\bsnm{Ommer}, \binits{B.}}:
\bctitle{Taming transformers for high-resolution image synthesis. 2021 ieee}.
In: \bbtitle{CVPR},
vol. \bseriesno{10}
(\byear{2020})
\end{bchapter}
\endbibitem

\bibitem[\protect\citeauthoryear{Ho and Salimans}{2022}]{ho2022classifier}
\begin{botherref}
\oauthor{\bsnm{Ho}, \binits{J.}},
\oauthor{\bsnm{Salimans}, \binits{T.}}:
Classifier-free diffusion guidance.
arXiv preprint arXiv:2207.12598
(2022)
\end{botherref}
\endbibitem

\bibitem[\protect\citeauthoryear{Grammatikopoulou
  et~al.}{2021}]{grammatikopoulou2021cadis}
\begin{barticle}
\bauthor{\bsnm{Grammatikopoulou}, \binits{M.}},
\bauthor{\bsnm{Flouty}, \binits{E.}},
\bauthor{\bsnm{Kadkhodamohammadi}, \binits{A.}},
\bauthor{\bsnm{Quellec}, \binits{G.}},
\bauthor{\bsnm{Chow}, \binits{A.}},
\bauthor{\bsnm{Nehme}, \binits{J.}},
\bauthor{\bsnm{Luengo}, \binits{I.}},
\bauthor{\bsnm{Stoyanov}, \binits{D.}}:
\batitle{Cadis: Cataract dataset for surgical rgb-image segmentation}.
\bjtitle{Medical Image Analysis}
\bvolume{71},
\bfpage{102053}
(\byear{2021})
\end{barticle}
\endbibitem

\bibitem[\protect\citeauthoryear{Allen and Vasavada}{2006}]{allen2006cataract}
\begin{barticle}
\bauthor{\bsnm{Allen}, \binits{D.}},
\bauthor{\bsnm{Vasavada}, \binits{A.}}:
\batitle{Cataract and surgery for cataract}.
\bjtitle{BMJ}
\bvolume{333}(\bissue{7559}),
\bfpage{128}--\blpage{132}
(\byear{2006})
\end{barticle}
\endbibitem

\bibitem[\protect\citeauthoryear{Bi{\'n}kowski
  et~al.}{2018}]{binkowski2018demystifying}
\begin{botherref}
\oauthor{\bsnm{Bi{\'n}kowski}, \binits{M.}},
\oauthor{\bsnm{Sutherland}, \binits{D.J.}},
\oauthor{\bsnm{Arbel}, \binits{M.}},
\oauthor{\bsnm{Gretton}, \binits{A.}}:
Demystifying mmd gans.
arXiv preprint arXiv:1801.01401
(2018)
\end{botherref}
\endbibitem

\bibitem[\protect\citeauthoryear{Zhang et~al.}{2018}]{zhang2018unreasonable}
\begin{bchapter}
\bauthor{\bsnm{Zhang}, \binits{R.}},
\bauthor{\bsnm{Isola}, \binits{P.}},
\bauthor{\bsnm{Efros}, \binits{A.A.}},
\bauthor{\bsnm{Shechtman}, \binits{E.}},
\bauthor{\bsnm{Wang}, \binits{O.}}:
\bctitle{The unreasonable effectiveness of deep features as a perceptual
  metric}.
In: \bbtitle{CVPR},
pp. \bfpage{586}--\blpage{595}
(\byear{2018})
\end{bchapter}
\endbibitem

\bibitem[\protect\citeauthoryear{He et~al.}{2017}]{he2017mask}
\begin{bchapter}
\bauthor{\bsnm{He}, \binits{K.}},
\bauthor{\bsnm{Gkioxari}, \binits{G.}},
\bauthor{\bsnm{Doll{\'a}r}, \binits{P.}},
\bauthor{\bsnm{Girshick}, \binits{R.}}:
\bctitle{Mask r-cnn}.
In: \bbtitle{ICCV},
pp. \bfpage{2961}--\blpage{2969}
(\byear{2017})
\end{bchapter}
\endbibitem

\end{thebibliography}

\newpage
\begin{appendices}

\section{Scene Graph Generation}
\label{sec:app_sg}

\begin{figure}[htbp]
    \centering
    \includegraphics[width=\linewidth]{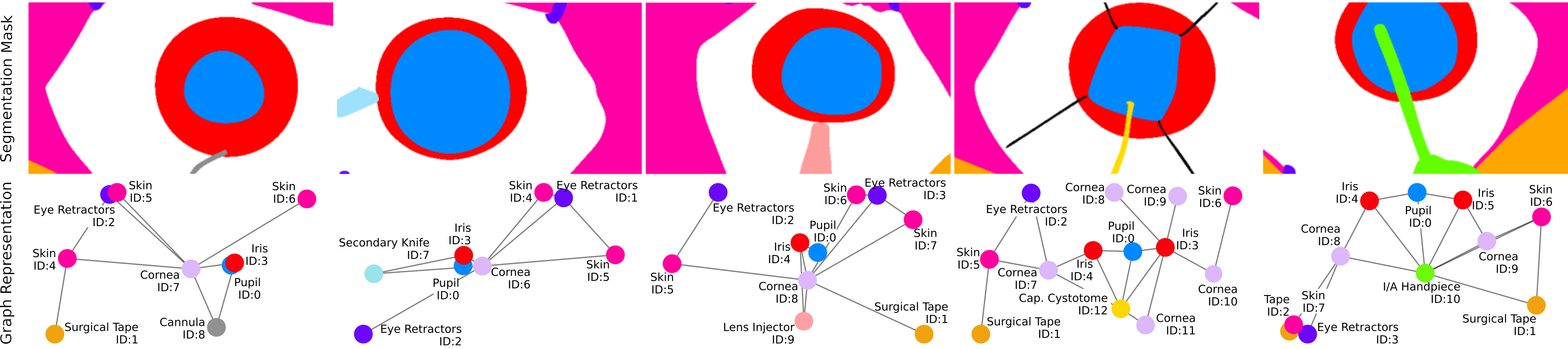}
    \caption{\textbf{Surgical Scene Graphs.} Each connected component in the segmentation mask is assigned a node. We store the component's class label, spatial spreading, and centroid coordinate within the node features. Visualized in the bottom row, the node positions correspond to these centroid coordinates. Edges between nodes indicate that components are spatially connected in the segmentation mask.}
    \label{fig:sg}
\end{figure}

We assume our dataset $\mathcal{D}$ consists of (image, mask) pairs $(x_\text{i},m_\text{i}) \in \mathcal{D}$. From each mask $m_\text{i}$, we build a Scene Graph $\mathcal{G}_\text{i}$ by extracting the masks’ connected components. We then compute the centroid of each component. Each node of the final graph represents an individual connected component and encodes the component’s \textit{$d$-dim} class vector, the 2-\textit{dim} spatial spreading of the component, and the component’s 2-\textit{dim} centroid coordinates. By explicitly encoding these values in the node features of the graph, we can precisely control the image synthesisation. In the graph’s edges, we encode the spatial relationship of nodes, for which we check if nodes’ components are touching spatially. Therefore, a SG for sample $i$ is defined by $\mathcal{G}_\text{i} = (\mathcal{V}_\text{i},\mathcal{E}_\text{i})$ where $\mathcal{V}_\text{i} \in \mathbb{R}^{n\times(d + 4)}$ are the node features, with $n$ being the number of nodes in the graph, and $\mathcal{E}_\text{i}$  is the set of undirected edges between connected nodes. Our surgical SG $\mathcal{G}_\text{i}$ yields a compact yet spatially and semantically accurate representation of a surgical image with a clear interpretation, as presented in Figure \ref{fig:sg}. Further, graph representations allow seamless modification of generative model conditioning by changing node features, which we demonstrate in Section \ref{sec:exp}. 

\section{Pre-Training Embedding Space Visualisation}
\label{sec:app_tsne}
\begin{figure}[htbp]
    \centering
    \includegraphics[width=1\textwidth]{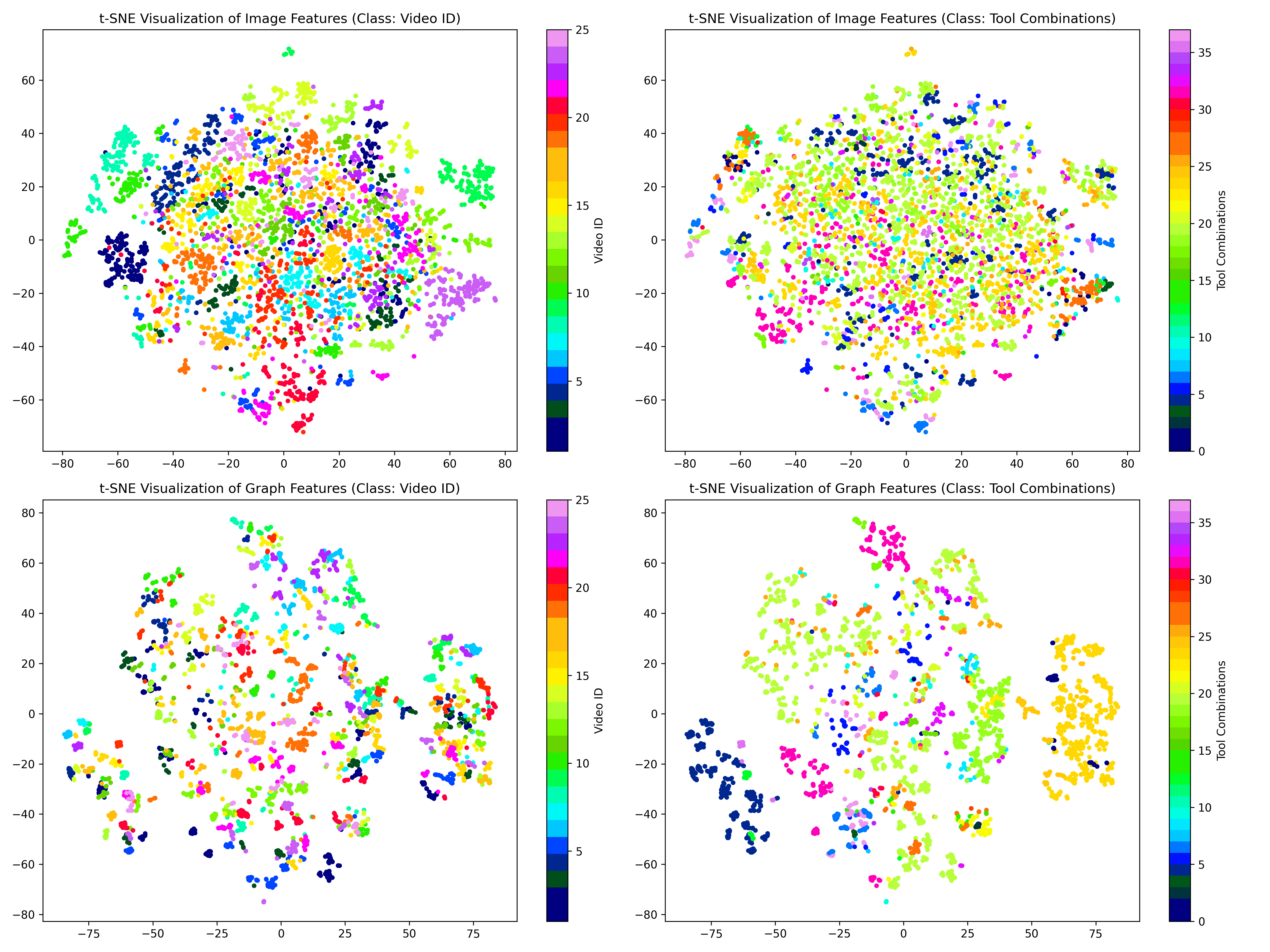}
    \caption{\textbf{t-SNE Embedding Space Visualisation.} }
    \label{fig:tsne_visualisation}
\end{figure}

Figure \ref{fig:tsne_visualisation} presents t-SNE visualizations of image and graph embeddings after the pre-training stage. The t-SNE plots in the first column are colour-coded by video ID, indicating that frames from the same video share the same colour. In contrast, the t-SNE plots in the second column are classified by the combinations of present tools.

These visualizations strongly support our claims. In the first column, when image features are classified by video ID, evident signs of clusterings can be seen, demonstrating that frames from the same video are grouped in the image embedding space. This clustering occurs regardless of the tools present in the frames, majorly due to shared anatomical features across frames. However, the top-right t-SNE plot shows little to no clustering when the embeddings are classified by tool combinations, reinforcing the observation that image embeddings fail to encode tool-specific information effectively. With pre-trained graph embeddings, clear clusters emerge, as shown in the bottom-right t-SNE plot. This indicates that frames with similar surgical tools and positions are close in the embedding space, irrespective of their video origin. On the other hand, when graph embeddings are classified by video ID, there is minimal clustering, as shown in the bottom-left t-SNE plot. Overall, this demonstrates that our pre-training method captures embeddings that are not dominated by video-specific information and clusters frames with similar surgical tools and positions in the embedding space.

\section{Expert Assessment Study - Feedback}
\label{sec:app_fb}
After participation, we asked participants to explain their reasoning for assigning low or high values to the realism and coherence of the generated images.
In summary, they mentioned that extreme changes to the input SG could be better reflected in the generated image. This primarily concerns the spatial positions of tools that are far from what the training dataset contains. The participants occasionally observed image parts that could be more realistic, e.g., off-looking anatomical textures, warped instruments, or a blurry or unrealistic interaction between tool and tissue. 
Additionally, they observed inconsistencies within the mini-batch of generated images. Individual images would sometimes show wrong or no tools.
Lastly, they also gave feedback on the graph representation, which they found easy to operate to condition the simulation. However, they wished for even more control, e.g., controlling the size of the pupil or the angle of the tools, and an even further simplified representation, e.g., not showing one node per component but rather per anatomy class.
Nonetheless, we received overall positive feedback for the general idea and the generated images' realism and coherence with the conditioning, reflected in the scores in Section \ref{sec:study}.

\section{Expert Assessment Study - GUI}
\label{sec:app_gui}
\begin{figure}[htbp]
    \centering
    \includegraphics[width=0.95\linewidth]{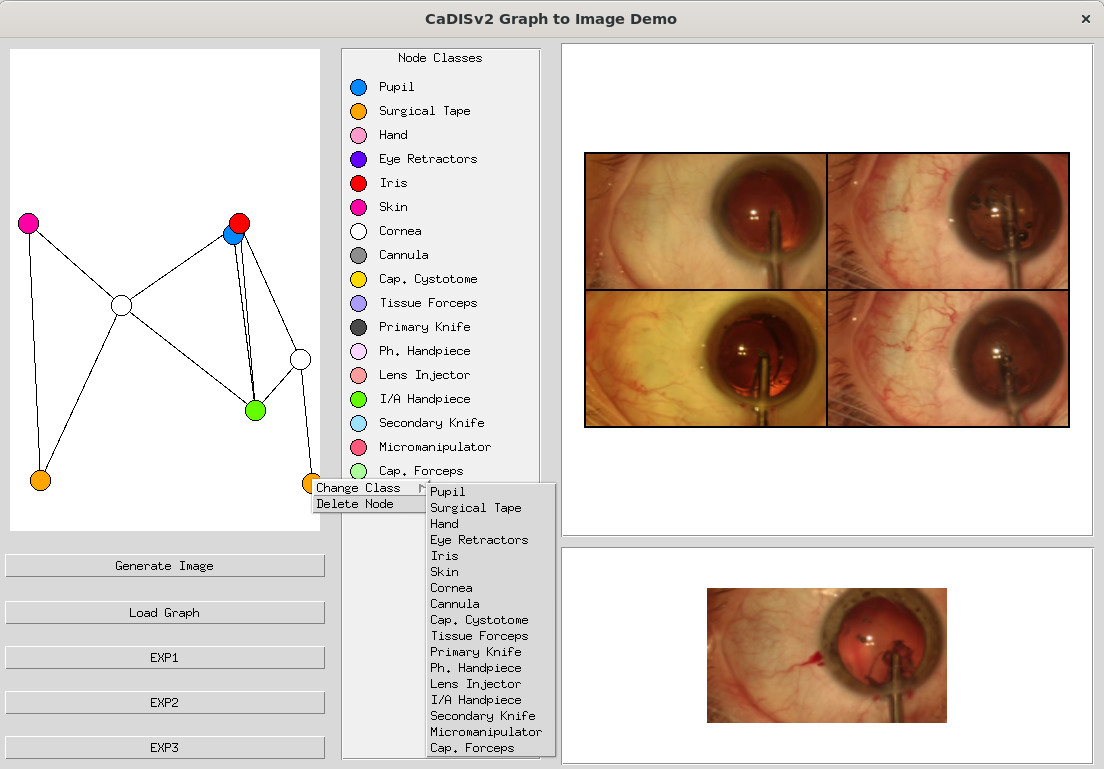}
    \caption{\textbf{Graphical User Interface for Expert Assessment.} In our GUI, participants can load ground truth (image, graph) pairs (bottom right and left). The GUI allows the following operations for modifying the graph: Nodes can be \emph{moved} spatially by drag-and-drop. They can be \emph{deleted} or \emph{changed into different classes} by right-clicking them. Further, \emph{new nodes can be added} by right-clicking the empty space in the SG canvas. Eventually, participants can generate a mini-batch of four images using the currently displayed SG (top right).}
    \label{fig:gui}
\end{figure}

This section presents the GUI used during our user study in Figure \ref{fig:gui}.

\section{Limitations \& Future Work}
\label{sec:app_lim}
In future work, we will primarily address the feedback from our user study, hence exploring ways to improve the usability of the GUI and SG representations. We also want to allow additional control over the pupil size and the tool angle. Since we see frame-based simulation only as a first step in building a surgical simulator, we will further explore sequence generation in the future. For example, sequential editing of an initial reference frame could substantially improve the temporal consistency of image information not encoded in the SG representation, such as textures or blood vessels. Additionally, we will explore other surgical domains in future research. We expect these directions to strengthen the possibilities for clinical translation significantly.

\end{appendices}

\end{document}